\documentclass[11pt,a4paper,table]{article}
\usepackage[hyperref]{emnlp2020}
\usepackage{times}
\usepackage{latexsym}

\usepackage{microtype}

\aclfinalcopy 

\usepackage{graphicx}
\usepackage{enumitem}
\usepackage{multirow}
\usepackage{subcaption}
\usepackage{float}
\usepackage{amssymb}
\usepackage{pifont}
\usepackage{chngcntr}
\usepackage[printwatermark]{xwatermark}

\title{Leveraging Multilingual News Websites for \\Building a Kurdish Parallel Corpus}

\author{Sina Ahmadi \\
  \normalsize{Insight Centre for Data Analytics}\\
  \normalsize{National University of Ireland Galway}\\
  \texttt{\small{ahmadi.sina@outlook.com}} \\\And 
  Hossein Hassani \\
  \normalsize{University of Kurdistan Hewlêr} \\
  \normalsize{Kurdistan Region, Iraq}   \\
  \texttt{\small{hosseinh@ukh.edu.krd}} \\\And
  Daban Q. Jaff \\
  \normalsize{Koya University}\\
  \normalsize{Kurdistan Region, Iraq}   \\
  \texttt{\small{daban.jaff@koyauniversity.org}} \\}

\date{}

\begin{document}
\maketitle
\begin{abstract}
    Machine translation has been a major motivation of development in natural language processing. Despite the burgeoning achievements in creating more efficient machine translation systems thanks to deep learning methods, parallel corpora have remained indispensable for progress in the field. In an attempt to create parallel corpora for the Kurdish language, in this paper, we describe our approach in retrieving potentially-alignable news articles from multi-language websites and manually align them across dialects and languages based on lexical similarity and transliteration of scripts. We present a corpus containing 12,327 translation pairs in the two major dialects of Kurdish, Sorani and Kurmanji. We also provide 1,797 and 650 translation pairs in English-Kurmanji and English-Sorani. The corpus is publicly available under the CC BY-NC-SA 4.0 license.\footnote{\url{https://github.com/KurdishBLARK/InterdialectCorpus}}
\end{abstract}

\section{Introduction}
\label{introduction}

For over half a century, machine translation has been one of the well-studied subjects in natural language processing (NLP) \cite{hutchins2005current,cheragui2012theoretical}. Although the operating principles of machine translation has been constantly improving from rule-based methods to statistical and neural network approaches, parallel corpora have remained essential components to efficiently address the complexity of human language in the translation task. A parallel corpus contains translation pairs in two languages or dialects that can be used for training translation models and learning the alignment of words and their placements within phrases. Creating such a resource is a tedious and time-consuming task that requires thorough linguist knowledge of the source and target languages. Oftentimes, lack of financial support further constrains the development of such resources for less-resourced languages, particularly Kurdish \cite{allah2012toward}. 

Multi-language news websites often provide similar content in different languages or dialects based on the same news source. Although the choice of the translators and editors determines how the original article is differently narrated in two different languages or dialects, such relevant news articles usually represent significant overlaps. Recently, parallel corpus filtering and alignment of crawled text from the web has gained more attention in the machine translation community \cite{koehn2018findings,sen2019parallel,koehn2019findings,steingrimsson2020effectively}.

In the same vein, we create a parallel corpus for the Kurdish language by collecting news articles from some of the multilingual Kurdish news websites. Relying on key elements of a news article, such as date of publication, topic and image URL, our approach filters articles at  document-level. Given the diversity of the alphabets in our case, i.e. Arabic-based Kurdish alphabet for content in Sorani and Latin-based alphabet for English and Kurmanji, we also use transliteration to calculate basic string similarities. The most similar headlines of the filtered documents are then provided to native annotators who verify the relatedness of the news articles. This way, we could collect 1,452 Sorani-Kurmanji, 282 English-Sorani and 277 English-Kurmanji articles. Following this step, the content of the relevant articles are automatically extracted and manually aligned at sentence level, yielding 12,327, 1,797 and 650 translation pairs in Sorani-Kurmanji, Sorani-English and Kurmanji-English.

The rest of the paper is organized as follows. We first provide a description of the previous work in the creation and alignment of parallel corpora and also present the available resources for Kurdish in Section \ref{relatedwork}. In Section \ref{kurdish_language}, we briefly describe some of the grammatical aspects of Kurdish and English which are important in translation. Section \ref{methodology} presents our approach on how the data is retrieved and aligned. Our parallel corpus is evaluated in Section \ref{evaluation}. Finally, the paper in concluded in Section \ref{conclusion}.

\section{Related Work}
\label{relatedwork}


During the early time of emergence of the Web contents, \citet{resnik2003} addressed and discussed the usage of the Web for developing parallel corpora. In the absence or limited availability of the digitized translated literature or other documents that usually could form the basis of parallel corpora, the Web content has become a significant resource for the development of the parallel corpora. Literature reports on the usage of the Web contents for the development of parallel corpora in the absence of available data in various cases, particularly for less-resourced languages \cite{morishita-etal-2020-jparacrawl,mubarak2020constructing,chiruzzo2020development}. For instance, \citet{inoue2018parallel} develop a parallel corpus for Arabic-Japanese based on news articles which is then manually aligned at the sentence level. Having said that, with the diversity of themes of the Web content, the representativeness of the developed corpus using this content could become an issue \cite{tadic2000building}. 

Regardless, the news content, whether online or paper-based has remained as one of the main sources for the parallel corpus development \cite{fry2005assembling,inoue2018parallel,mino2020content,toral-2014-tlaxcala}.  

Regarding the Kurdish language, efforts has increased recently to create language resources, such as lexicographical resources \cite{ahmadi2019lex}, monolingual corpora \cite{esmaili2013building,abdulrahman2019ktc} and even a folkloric corpus \cite{ahmadi2020folklyrics}. These have improved the situation that was reported by \citet{hassani2018blark}. Moreover, the construction of inter-dialectal resources for Kurdish has been of interest previously. \citet{hassani2017kurdish} studies the application of word-by-word translation for translating Kurmanji to Sorani using a bi-dialectal dictionary. The study aims to evaluate the efficiency of the method in the absence of parallel corpora. Although the experiments show reasonable outcome, the study reports unnaturalness in the translation.

There are fewer resources that include Kurdish parallel texts. The Tanzil corpus\footnote{\url{http://tanzil.net}} which is a compilation of Quran translations, various Bible translations\footnote{\url{http://ibtrussia.org/en/}}, the TED corpus\footnote{\url{https://wit3.fbk.eu/}} \cite{cettolo2012wit3} and the KurdNet--the Kurdish WordNet \cite{aliabadi2014towards} provide translations in Sorani Kurdish. In addition to Bible translations, Kurmanji has received further attention in the machine translation realm. For instance, \citet{ataman2018bianet} reports on the creation of one parallel corpus for Kurmanji-Turkish-English. Moreover, Google Translate\footnote{\url{https://translate.google.com/}}, the Google translation service, provides Kurmanji in the list of its languages. Although the resources are not openly available, we believe that crowd-sourcing projects contribute to such projects. 

In order to create a parallel corpus for the Sorani-Kurmanji dialects of Kurdish and also, as a preliminary effort to create Sorani-English and Kurmanji-English parallel corpora, we report our endeavour to create parallel corpora for Kurdish based on the content of Kurdish News websites. 


\section{Kurdish Language}
\label{kurdish_language}

\begin{table*}[!htbp]
\begin{subfigure}[b]{1\textwidth}
    \centering
\includegraphics{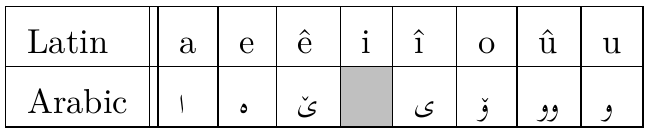}
    \caption{Vowels}
    \label{fig_alphabets_cons}
\end{subfigure}

\begin{subfigure}[b]{1\textwidth}
    \centering
    \includegraphics[scale=0.95]{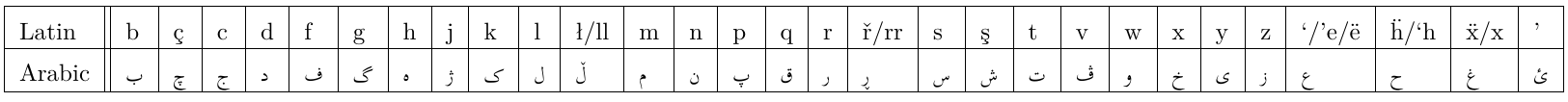}
    \caption{Consonants}
    \label{fig_alphabets_vow}
\end{subfigure}
\caption{A comparison of the Arabic and Latin based alphabets of Kurdish. Variations are specified with "/"}
\label{fig_alphabet_comp}
\end{table*}

\subsection{Alphabets and Dialects}

Some scholars categorize Kurdish as a dialect continuum for which language intelligibility varies from region to region \cite{haig2002kurdish}. Generally, Kurdish is believed to have three main dialects as Northern Kurdish (Kurmanji), Central Kurdish (Sorani) and Southern Kurdish \cite{matras2017}. These three dialects are spoken by 20-30 million speakers in the Kurdish regions of Iraq, Iran, Turkey and Syria \cite{ahmadi2019lex}. While many multi-dialect languages, such as Arabic or Chinese, exist in which one could find mutually unintelligible dialects, they usually have a standard form that regulates the communication among the speakers. Regarding Kurdish, although the standardization of the language, both in written and spoken forms, has been widely discussed, there is still no consensus among scholars and also the speakers \cite{khalid2015kurdish}. As a result, the language is written in many scripts, mainly Arabic-based and Latin-based, and each dialect is used as distinct languages in the media \cite{hassani2017kurdish,tavadze2019spreading}. Table \ref{fig_alphabet_comp} provides the alphabets used for writing Kurdish in a comparative way.

\begin{table*}[h]
\centering
\resizebox{1\linewidth}{!}{%
\begin{tabular}{|p{1.5cm}|l|p{2.5cm}|p{1.7cm}|p{3.1cm}|p{3.7cm}|}
\hline
Language          & Word order      &  Passive    & Gender & Case & Alignment\\ \hline\hline
Kurmanji Kurdish  & S-O-V      &  periphrastic with \textit{hatin} (to come) ~\cite{thackston2006kurmanji} &  feminine, masculine ~\cite{thackston2006kurmanji} & nominative, oblique, Izafa, vocative ~\cite{thackston2006kurmanji} & nominative–accusative, only in past transitive ergative–absolutive ~\cite{matras1997clause}\\ \hline
Sorani Kurdish    & S-O-V      &  morphological ~\cite{thackston2006sorani} & no gender ~\cite{thackston2006sorani} & nominative, locative, vocative ~\cite{mccarus2007kurdish} & nominative–accusative, only in past transitive ergative–absolutive ~\cite{karimi2014syntax}\\ \hline \hline
English          & S-V-O       &  periphrastic & no gender  & nominative, oblique, genitive only for personal pronouns & nominative–accusative \\ \hline
\end{tabular}
}
\caption{A comparison of the Sorani and Kurmanji dialects of Kurdish with English}
\label{tab_language_comparison}
\end{table*}

\begin{figure*}[t]
    \centering
\includegraphics[width=\linewidth]{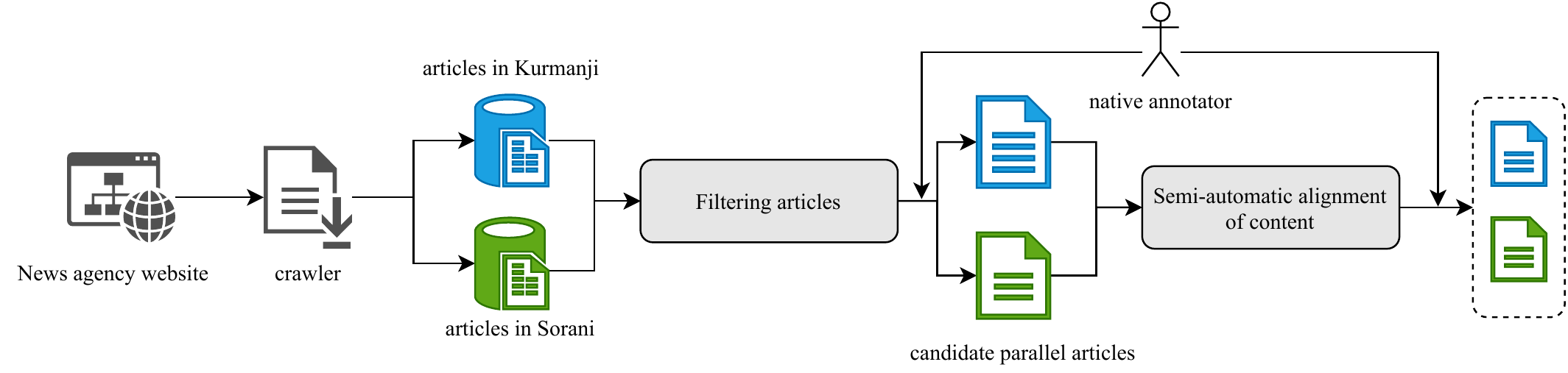}
    \caption{Our approach to automatically retrieve identical news articles}
    \label{fig_approach}
\end{figure*}

\subsection{Vocabulary}

The lexical diversity and richness of Kurdish has been previously attested by many lexicographers \cite{hejar1991,chyet2003kurdish,bedirxan2009ferheng,uok2012,uok2018}. This diversity is to such an extent that the vocabulary may vary from one village to another. Moreover, being in touch with many regional languages, especially Arabic, Persian, Turkish and Armenian, and local languages, particularly Zazaki and Gorani, almost all Kurdish dialects have entered many lexical borrowings into the language as well \cite{chyet2020ferhenga}. Having an oral tradition in narrating poetry and prose, the oral literature has been considered as a source of vocabulary by lexicographers \cite{ahmadi2020folklyrics}. In addition, there is an ongoing struggle to develop modern technical terminologies for the language.

Regarding Kurdish lexicographic resources, \cite{ahmadi2019lex} survey the current state of Kurdish lexicography and state that despite the scarcity of resources in electronic forms for Kurdish, there are over 71 dictionaries and terminological resources for Kurdish which are not all recto-digitized.

\subsection{Grammar}

Despite the lexical similarity between the dialects of Kurdish, there are differences when it comes to grammar, particularly due to morphological constructions. Sorani tends to have a more complex morphological construction while Kurmanji is less inflected. For instance, passive voice in Sorani is derived from the transitive verbs, while in Kurmanji, passive voice has a simpler construction where a compound is created by adding the auxiliary verb \textit{hatin} `to come' to the transitive verb without any major morphological modification \cite{thackston2006sorani}. In addition, Sorani has a full article marking system where nouns are marked as definite, indefinite, demonstrative in singular and plural forms while articles in Kurmanji are marked only in definite and demonstrative cases \cite{jugel2014linguistic}.

Regarding grammatical cases, unlike Sorani and English, Kurmanji has two grammatical genders, i.e. feminine and masculine, which implies a grammatical agreement particularly in \textit{Izafe} (also known as \textit{Ezafe}) constructions \cite{samvelian:halshs-00673182}. The Izafa construction refers to the usage of a grammatical particle to form noun phrases or adjective phrases. This grammatical particle in Kurmanji and Sorani are respectively \textit{-ê, -ekî, -a, -eke, -ên} and \textit{-î, -e} \cite{thackston2006kurmanji,salehi2018constraints}. Although in the adjective phrases, the particle is not translated, e.g. \textit{xanîy\textbf{ê} biçûk} ``the small house'', in the noun phrases it is usually translated as `of', e.g. \textit{xanîyê wî mirovî} ``the house of that man''. 

Table \ref{tab_language_comparison} provides some of the major grammatical characteristics of Kurmanji, Sorani and English. Both Kurdish dialects have a subject-object-verb alignment for present tenses and intransitive verbs, and an agent-object-verb alignment for transitive verbs in the past tense. The morphosyntactic property of agreement of the subject of intransitive verbs as the object (patient) of transitive verbs in the past tenses is known as ergativity and also exists in Kurdish \cite{karimi2014syntax}. Unlike Kurmanji Kurdish which uses oblique case of pronouns for this purpose, Sorani Kurdish only uses different pronominal clitics to demonstrate such an alignment \cite{esmaili2013sorani}.

It is worth mentioning that variations exist among Sorani subdialects, particularly the dialects which are categorized as Northern Sorani in \cite{matras2017} which take use of oblique cases and grammatical gender to some extent.

\section{Methodology}
\label{methodology}

Multilingual news websites contain a large number of articles in various languages which can be considered a potentially parallel corpus. However, among the major Kurdish news agencies, listed in Table \ref{tab_list_of_news_agencies}, none of them explicitly link identical articles across languages, e.g. by using reference keys or identical URL schema or news code. Moreover, only a few of them provide the same content in various languages. For instance, the English articles on BasNews are different in content and topic in comparison to the Kurdish ones.

In this section, we describe our approach which is illustrated in Figure \ref{fig_approach}, to create a parallel corpus of Sorani, Kurmanji and English. We refer to these three as languages for ease of reference.

\subsection{Data Crawling}

As the first step, we crawl the content of news websites. Our selection criteria are the editorial quality of the articles, accessibility of the data to be automatically scraped and more importantly, multilingualism. Therefore, we selected Firat News Agency (ANF), BasNew (BN) and KurdPa (KP). Despite the remarkable size of articles published on Rûdaw and Kurdistan 24, we could not include those websites due to crawling restrictions. Moreover, our findings regarding the alignment of Voice of America was not satisfying due to sparsity of topics across languages.

Once the news articles are crawled, we clean the HTML files and extract the following information from each page:

\begin{itemize}[nolistsep]
    \item \texttt{tag}: a list of the tags used for identifying the article. For these purpose, \texttt{bashakan}, \texttt{cat-links}, \texttt{keywords} tags were originally used in BN, KP and VOA, respectively. In the case of ANF, we used the page hyperlink structure to extract the topic and used it as a tag.
    \item \texttt{original\_link}: the original link to the article on the website
    \item \texttt{dialect}: the dialect of the article retrieved using the link schema, usually \texttt{so} for Sorani and \texttt{ku} for Kurmanji
    \item \texttt{entry-title}: the news headline
    \item \texttt{entry-lead}: the news sub-headline, if provided
    \item \texttt{date}: the publication date of the article. We unified all the date formats based on the Gregorian calendar given the variety of calendars, e.g. Kurdish or Persian calendars
    \item \texttt{entry-content}: a list containing paragraphs, i.e. \texttt{<p>}, provided in the content of each news article. The content of our target websites are originally marked with the \texttt{<entry-content>} tag.
    \item \texttt{imgs}: Assuming that relevant news articles link to the same multimedia content with the same hyperlink, we retrieve the hyperlinks associated to the \texttt{<img>} tags within the body of the article
\end{itemize}

In addition to the HTML tags, in some cases we could use JSON-LD and the meta tags, i.e. \texttt{<meta>}, to retrieve further instances. Ultimately, the news articles of each website are normalized and categorized by dialect and language in JSON format.

\begin{table*}[h]
\centering
\scalebox{1}{
\begin{tabular}{|l|p{8cm}|}
\hline
agency       & languages           \\ \hline \hline
\href{https://www.rudaw.net}{Rûdaw}    &     Sorani, Kurmanji, English, Arabic, Turkish  \\ \hline 
\href{https://www.dengiamerika.com}{Voice of America} & Sorani, Kurmanji, English, Turkish and many more  \\ \hline 
\href{https://www.kurdistan24.net}{Kurdistan24}         &  Sorani, Kurmanji, English, Arabic, Turkish, Persian  \\ \hline 
\href{https://www.knnc.net}{KNN}  & Sorani, English, Arabic \\ \hline 
\href{https://anfsorani.com}{Firat News Agency} & Sorani, Kurmanji, Zazaki, Gorani, English, Arabic, Turkish, Persian, German, Russian, Spanish  \\ \hline 
\href{http://bianet.org}{Bianet} & Kurmanji, English, Turkish \\ \hline 
\href{http://www.basnews.com}{BasNews}  &  Sorani, Kurmanji, English, Arabic, Turkish, Persian \\ \hline 
\href{https://kurdpa.net}{KurdPa} &  Sorani, Kurmanji, English, Persian \\ \hline 
\href{http://www.gulan-media.com}{Gulan Media}  & Sorani, Kurmanji, English, Arabic  \\ \hline 
\href{https://www.nrttv.com}{NRT} & Sorani, English, Arabic \\ \hline 
\href{https://sahartv.ir}{SaharTV} & Sorani, Kurmanji, English, Persian \\ \hline 
\end{tabular}
}
\caption{List of news agencies providing content in Kurdish and their content management status}
\label{tab_list_of_news_agencies}
\end{table*}

\subsection{Corpus Filtering}

Given two sets of articles of the same news website in two languages, we consider two articles alignable if they, at least, have one common tag and identical publication dates with the exact month and year. Intuitively speaking, two articles published in two different years with two different tags (topics) are less probable to contain the same content. In addition to this, we also use \texttt{<imgs>} to filter the articles in such a way that if two articles have linked to the same image, we consider them potentially alignable. 

Moreover, as several news articles could be published with the same tags within the same date range, we further filter out the candidate articles by comparing the headlines. To do so, we calculate the similarity of the headlines based on the a simple string sequence matching scorer. In the case of Sorani, as it is written in the Arabic-based alphabet, we first transliterate the Sorani text, using \textsc{Wergor} \cite{ahmadi2019wergor}, into the Latin-based script which is used for Kurmanji and English. 

As the final step, among the candidate headlines, we retrieve the top five most similar headlines in the other language. These headlines are then provided in spreadsheets to native annotators who determine if two headlines correspond to the same news content using a drop-down list. If two headlines are literal translations and refer to the same content, they are specified as \textit{equivalent}. However, this is not always the case as some headlines are paraphrases and rewritten in such a way that they attract the readers' attention. In such cases where two headlines refer to the same content but are not literal translations, they are annotated as \textit{possible}. Although we do not consider such headlines as a translation pair, they are essential to retrieve relevant contents. In the cases where the headlines do not provide sufficient information to decide their relatedness, annotators are asked to check the crawled data in the two languages manually. Figure \ref{fig_spreadsheet} in Appendix \ref{sec_appendix} illustrates an annotation example in Kurmanji and English. 

\begin{table*}[h]
\centering
\resizebox{2\columnwidth}{!}{%
\begin{tabular}{|l|l|l|p{3cm}|p{3cm}|p{3cm}|}
\hline
News agency & \multicolumn{2}{l|}{Articles}              & \texttt{kmr}-\texttt{eng}                  & \texttt{ckb}-\texttt{eng}                  & \texttt{ckb}-\texttt{kmr}          \\ \hline \hline
\multirow{4}{*}{ANF} & \multicolumn{2}{l|}{\# crawled} & \multicolumn{3}{l|}{3026 (\texttt{kmr}) - 2432 (\texttt{ckb}) - 2937 (\texttt{eng})} \\ \cline{2-6} 
&   & \# headlines & 203 (132 \texttt{\textless{}img\textgreater{}})         & 212 (70 \texttt{\textless{}img\textgreater{}})          & 773 (507 \texttt{\textless{}img\textgreater{}}) \\ \cline{3-6} 
&    & \# sentences &  1,466    &   381    & 11,278 \\ \cline{3-6} 
& \multirow{-3}{*}{Retrieved} & \# tokens    &  \mbox{26,591 (\texttt{kmr})} \mbox{26,832  (\texttt{eng})}  &  \mbox{4,804  (\texttt{ckb})}  \mbox{5,130 (\texttt{eng})}   &  \mbox{179,225  (\texttt{ckb})}  \mbox{201,758  (\texttt{kmr})}   \\ \cline{1-6}
\multirow{4}{*}{KP}  & \multicolumn{2}{l|}{\# crawled} & \multicolumn{3}{l|}{663 (\texttt{kmr}) - 1,281 (\texttt{ckb}) - 566 (\texttt{eng})}   \\ \cline{2-6} 
&        & \# headlines & 74 (74 \texttt{\textless{}img\textgreater{}})       & 80 (70 \texttt{\textless{}img\textgreater{}})           & 135 (101 \texttt{\textless{}img\textgreater{}}) \\ \cline{3-6} 
&      & \# sentences &      331     &  269       & 920    \\ \cline{3-6} 
& \multirow{-3}{*}{Retrieved} & \# tokens    &          \mbox{8,310 (\texttt{kmr})} \quad \mbox{7,375 (\texttt{eng})}      &        \mbox{6,416 (\texttt{ckb})}  \quad  \mbox{6,725 (\texttt{eng})}   & \mbox{23,732 (\texttt{ckb})}  \mbox{27,550 (\texttt{kmr})}                \\ \cline{1-6} 
\multirow{4}{*}{BS}  & \multicolumn{2}{l|}{\# crawled} & \multicolumn{3}{l|}{1,177 (\texttt{kmr}) - 1,277 (\texttt{ckb}) - 701 (\texttt{eng})}  \\ \cline{2-6}
&    & \# headlines & \cellcolor[HTML]{C0C0C0} & \cellcolor[HTML]{C0C0C0} &        32    \\ \cline{3-6} 
&      & \# sentences & \cellcolor[HTML]{C0C0C0} & \cellcolor[HTML]{C0C0C0} &    129      \\ \cline{3-6} 
& \multirow{-3}{*}{Retrieved}  & \# tokens & \cellcolor[HTML]{C0C0C0} & \cellcolor[HTML]{C0C0C0} & \mbox{2,425 (\texttt{ckb})} \mbox{2,251 (\texttt{kmr})} \\ \hline \hline
\multirow{3}{*}{All}  &       & \# headlines & 281 & 292 &   1,037               \\ \cline{3-6} 
&       & \# \textbf{sentences} & 1,797 & 650 &     12,327             \\ \cline{3-6} 
& \multirow{-3}{*}{Retrieved}  & \# \textbf{tokens} & \mbox{34,901 (\texttt{kmr})} \quad  \mbox{34,207 (\texttt{eng})} & \mbox{11,220 (\texttt{ckb})} \quad \quad \mbox{11,855 (\texttt{eng})}  & \mbox{205,382 (\texttt{ckb})}  \mbox{231,559 (\texttt{kmr})}  \\ \cline{1-6} 
\end{tabular}
}
\caption{Statistics of the Kurmanji (\texttt{kmr}), Sorani (\texttt{ckb}) and English (\texttt{eng}) articles used to create our parallel corpus. \texttt{\textless{}img\textgreater{}} refers to the articles retrieved through the image URLs in the HTML source code}
\label{tab_basic_statistics}
\end{table*}

\subsection{Content Alignment}

As the result of the previous steps, a list of the alignable articles of the same news website in two languages is available. Using the aligned headlines, we collect their contents, i.e. the content of \texttt{<entry-lead>} and \texttt{<entry-content>}, and provide them in two separate files where paragraphs and articles are respectively separated by one and two new lines. These files are then provided to the native annotators who extract parallel sentences and phrases in the two languages using \textsc{InterText} \cite{VONDRICKA14}. \textsc{InterText}\footnote{\url{https://wanthalf.saga.cz/intertext}} is an editor for aligning parallel texts and provides a wide range of editing functions such as merge, split and positioning. 

In the manual alignment task, we extract translation pairs based on the following guidelines:

\begin{enumerate}[noitemsep]
    \item the length of the sentences or phrases should be within a reasonable range. If too long, they are to be split into smaller phrases
    \item idiomatic translations are validated as long as they do not add to the size of the sentence significantly
    \item if the translation of a sentence is provided in many separate sentences or phrases, the annotator is allowed to merge the sentences to create a valid translation pair
    \item if two sentences can be validated with slight modifications, such as punctuation marks or digits, the annotator is allowed to edit the content
\end{enumerate}

\section{Evaluation}
\label{evaluation}

Table \ref{tab_basic_statistics} presents basic statistics of the corpus where the whole number of crawled articles and the number of retrieved articles among them are provided. We also specify the number of articles that are retrieved using multimedia hyperlinks using \texttt{\textless{}img\textgreater{}}.

In all the translation pairs, 17 to 20 tokens are on average present in each sentence. In contrast, the average number of tokens in Tanzil, TED and KurdNet corpora is respectively around 25, 70 and 6. As such, we believe that our resources are comparatively better when it comes to automatic alignment.

In addition to the basic statistics, we used Moses \cite{koehn2007moses} to test and evaluate the usage of the corpus in the statistical machine translation. We divided the corpus into two sets, 90\% as a training set and 10\% as a test set. The training set received a higher percentage because of the relatively small size of our corpora. The sets were selected randomly. We prepared the random selection scripts in a way that the whole experiment is reproducible. We trained Moses according to its recommended procedures. We also tested the accuracy of the system based on the Moses guideline that provides the BLEU \cite{papineni2002bleu} evaluation based on the test set. Table \ref{tab_smt_baseline} presents the results of BLEU scores for the Sorani-English, Kurmanji-English, and Sorani-Kurmanji data.

\begin{table}[H]
\centering
\begin{tabular}{|l|l|}
\hline
Baseline system  & BLEU\\ \hline\hline
Sorani-Kurmanji &     17.08    \\ \hline
Sorani-English   &  17.74      \\ \hline
Kurmanji-English  &   11.06 \\ \hline
\end{tabular}
\caption{Results of a baseline statistical machine translation system trained on our parallel corpus}
\label{tab_smt_baseline}
\end{table}

\section{Conclusion and Future Work}
\label{conclusion}

In this paper, we report our efforts in creating a parallel corpus for the Kurdish language, as a less-resourced language. Given that manual translation is an expensive and tedious task, we used the content of multilingual Kurdish news websites to extract potentially-alignable Sorani, Kurmanji and English sentences in a semi-automatic manner. The candidate sentences are then provided to native speakers to validate if they are translation pairs. This way, the task of translation is carried out as an annotation task. Our corpus contains 12,327 Sorani-Kurmanji, 1,797 Kurmanji-English and 650 Sorani-English translation pairs.

As the material for machine translation, we believe that our resource can pave the way for further developments in Kurdish machine translation. In order to facilitate the alignment of the news articles, we also propose that a referencing mechanism be embedded within each news article so that corresponding texts could be linked more easily in the future. We would also like to suggest our approach to further extend the current corpus or create new corpora for the other dialects of Kurdish.



\bibliographystyle{acl_natbib}
\bibliography{references}

\onecolumn
\appendix
\section{Appendix}
\label{sec_appendix}

\counterwithin{figure}{section}

\begin{figure}[H]
    \centering
    \includegraphics[width=1\textwidth]{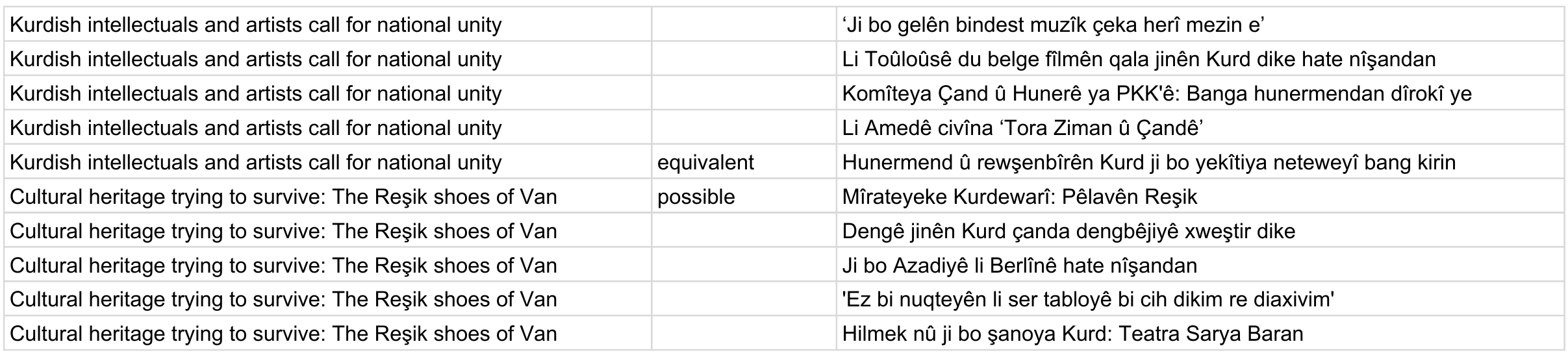}
    \caption{An example of the alignment of headlines. For each headline in English (left column), the five most similar headlines among the filtered Kurmanji headlines are provided. Using the drop-down list in the middle column, the annotator determines if two headlines are literal translations by selecting \textit{equivalent} or if they are not literal translation but correspond to each other by selecting \textit{possible}}
    \label{fig_spreadsheet}
\end{figure}

\begin{figure}[H]
    \centering
    \includegraphics[width=1\textwidth]{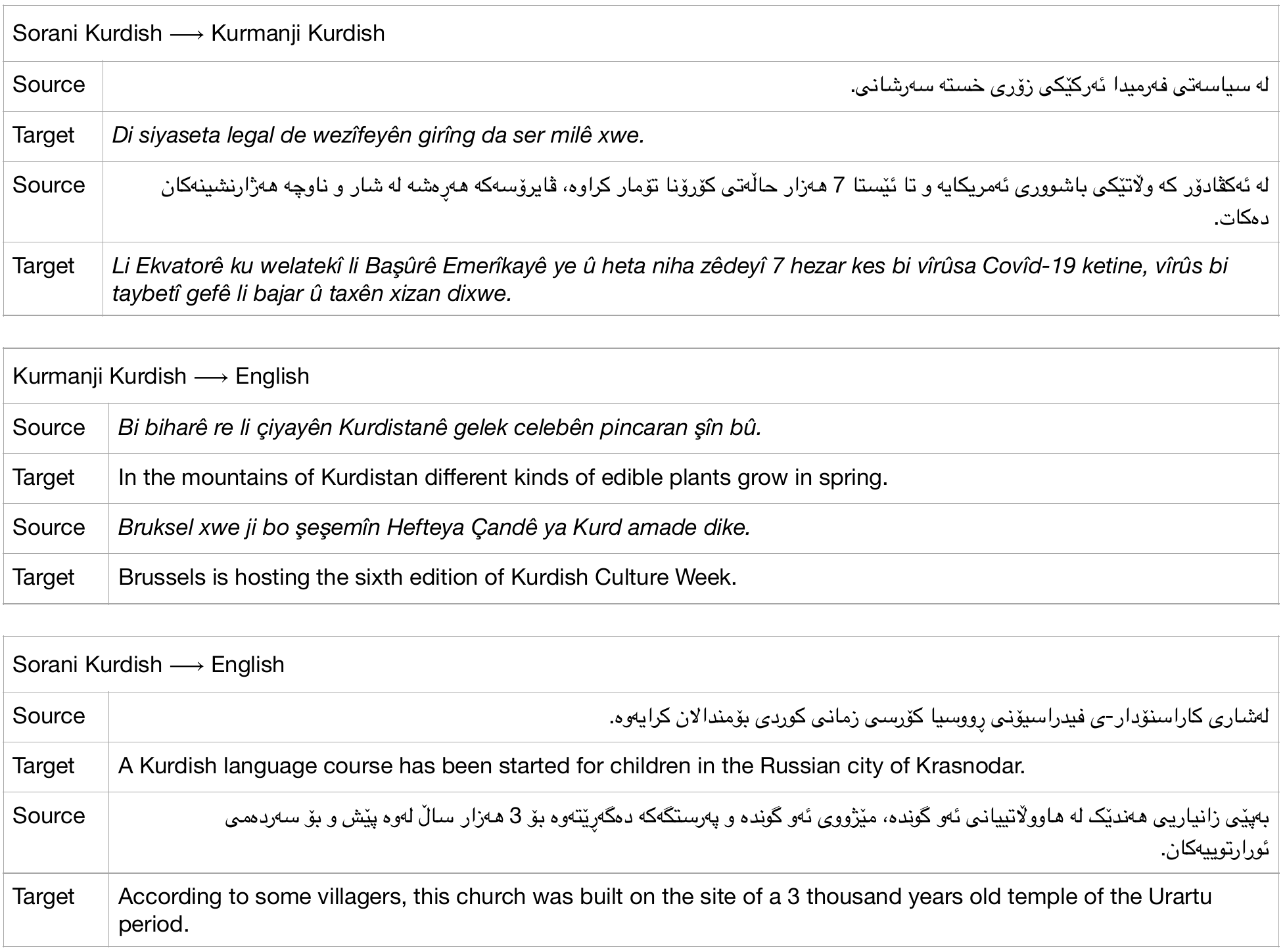}
    \caption{Examples of good translation pairs in our corpus}
    \label{fig_good_examples}
\end{figure}

\end{document}